\documentclass[letterpaper, 10 pt, conference]{ieeeconf}  

\usepackage{graphicx}
\usepackage{multirow}
\usepackage{xcolor}
\usepackage{algorithm}
\usepackage{algpseudocode}
\usepackage{mathtools}

\usepackage{hyperref}  
\usepackage{subcaption}
\usepackage{amsmath}
\usepackage{booktabs}
\usepackage{multicol}
\usepackage{amssymb}
\usepackage[colorinlistoftodos,prependcaption,textsize=tiny]{todonotes}
\usepackage{cite}
\usepackage{caption}
\graphicspath{ {./} }

\usepackage[a-1b]{pdfx}

\makeatletter
\newcommand\fs@spaceruled{\def\@fs@cfont{\bfseries}\let\@fs@capt\floatc@ruled
  \def\@fs@pre{\vspace{0.5\baselineskip}\hrule height.8pt depth0pt \kern2pt}%
  \def\@fs@post{\kern2pt\hrule\vspace{-0.95\baselineskip}}
  \def\@fs@mid{\kern2pt\hrule\kern2pt}%
  \let\@fs@iftopcapt\iftrue}
\makeatother

\IEEEoverridecommandlockouts                              
\title{\LARGE \bf
GS-NBV: a Geometry-based, Semantics-aware Viewpoint Planning Algorithm for Avocado Harvesting under Occlusions}

\author{Xiao'ao Song$^{1,2}$ and Konstantinos Karydis$^{1}$ 
\thanks{$^{1}$Autonomous Robots and Control Systems Lab, University of California, Riverside. Email: 
 karydis@ucr.edu.}%
\thanks{$^{2}$Perception, Robotics, AI, Sensing Lab, University of Colorado, Boulder. Email:xiaoao.song@colorado.edu.}%
\thanks{We gratefully acknowledge the support of NSF \#CMMI-2326309, USDA-NIFA \#2021-67022-33453, and The University of California under grant UC-MRPI M21PR3417. 
Any opinions, findings, and conclusions or recommendations expressed in this material are those of the authors and do not necessarily reflect the views of the funding agencies.}
}

\begin{document}

\maketitle
\thispagestyle{empty}
\pagestyle{empty}

\begin{abstract}
Efficient identification of picking points is critical for automated fruit harvesting. Avocados present unique challenges owing to their irregular shape, weight, and less-structured growing environments, which require specific viewpoints for successful harvesting. We propose a geometry-based, semantics-aware viewpoint-planning algorithm to address these challenges. The planning process involves three key steps: viewpoint sampling, evaluation, and execution. Starting from a partially occluded view, the system first detects the fruit, then leverages geometric information to constrain the viewpoint search space to a 1D circle, and uniformly samples four points to balance the efficiency and exploration. A new picking score metric is introduced to evaluate the viewpoint suitability and guide the camera to the next-best view. We validate our method through simulation against two state-of-the-art algorithms. Results show a 100\% success rate in two case studies with significant occlusions, demonstrating the efficiency and robustness of our approach. Our code is available at \href{https://github.com/lineojcd/GSNBV}{https://github.com/lineojcd/GSNBV}.
\end{abstract}

\section{Introduction}
Increasing labor costs and shortages pose significant challenges to the fruit harvesting industry~\cite{valle2014australian}.
Robot-assisted harvesting can offer a solution to enhance efficiency and complement human labor~\cite{kondo2011agricultural}. 
To be effective, efficient automated picking algorithms must be developed. 

Fruit-picking algorithms integrate perception to detect the target fruit (and/or relevant parts such as the peduncle) and motion planning to guide the robot's end-effector to specific locations (picking points). 
Because the picking-point distribution varies across fruit types, picking algorithms and end-effectors must be tailored accordingly. 
Fruits with relatively smooth surfaces and mostly spherical shapes (e.g., apples and oranges) can afford picking points distributed over the entire visible fruit surface. 
This facilitates end-effector design (e.g., retrieval by directly contacting the fruit and using vacuum-powered suction cups~\cite{hemming2014robot, bontsema2014crops, hayashi2010evaluation}) and may reduce algorithmic complexity (e.g., requiring only the fruit's 3D position to complete the harvesting task~\cite{li2022occluded}). 
More delicate and softer fruits (e.g., tomatoes and blueberries) can be picked up by wrapping pneumatically actuated deformable fingers around them~\cite{navas2024soft}. 
In this case, the picking points are still distributed over almost the entire fruit surface.
A different harvesting means for some fruits/vegetables growing in clusters (e.g., grapes~\cite{jiang2022development}, litchi~\cite{zhuang2019computer}) or along long stems (e.g., peppers~\cite{lehnert2017autonomous}, strawberries~\cite{yu2019fruit}) is to cut the peduncle and retrieve them. 
In these cases, obtaining the 3D position of the fruit or peduncle is sufficient to complete the harvesting task. 

\begin{figure}[!t]
\vspace{6pt}
    \centering
    \includegraphics[trim={0cm 0cm 0cm 0cm},clip,width=0.425\textwidth]{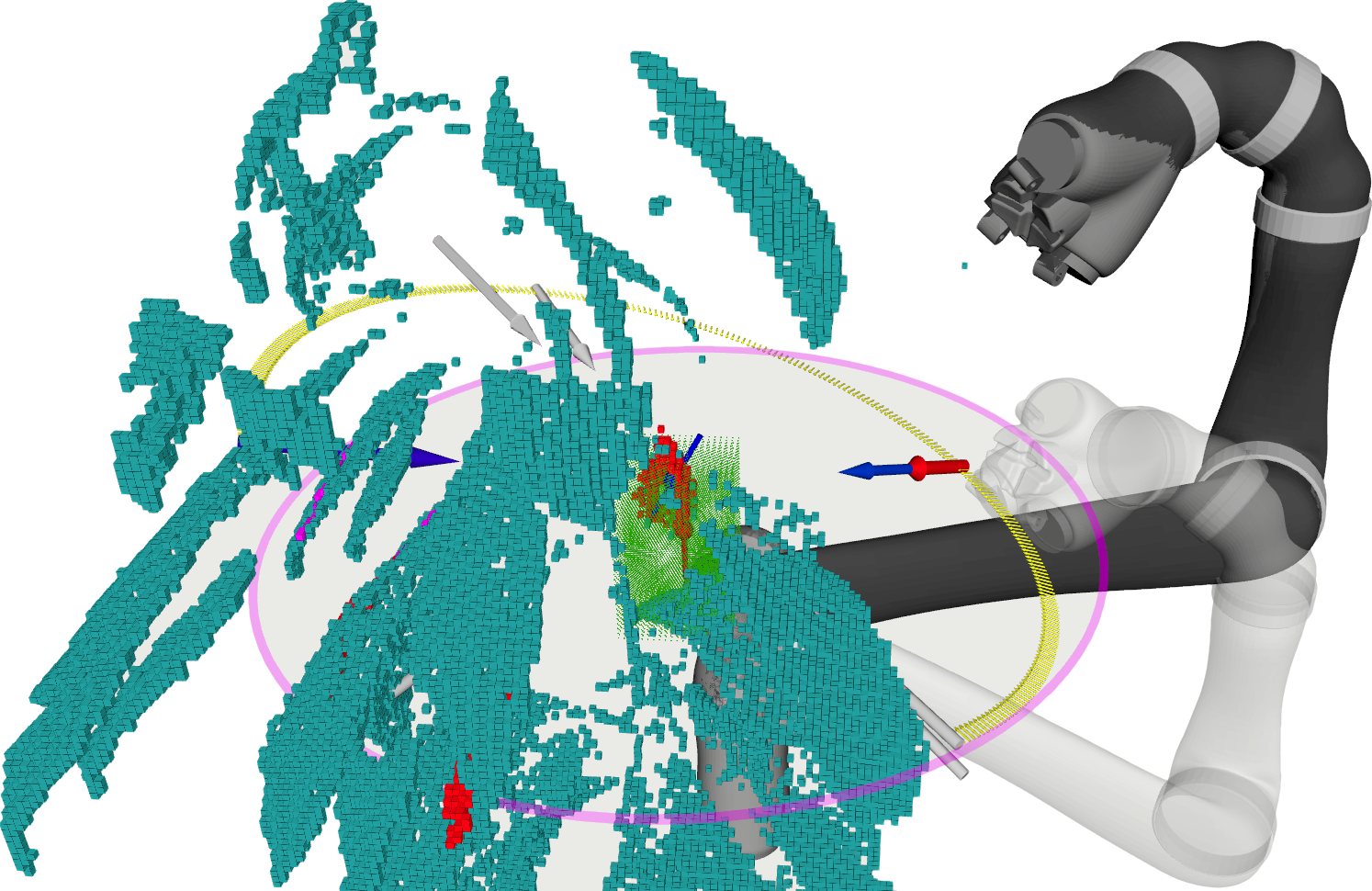}
    \vspace{0pt}
    \caption{
    GS-NBV estimates a picking ring (yellow circle; ground truth in magenta) around the avocado (in red) and moves the sensor along it to a pose that minimizes occlusion and is optimal for avocado grasping. 
    Sampled viewpoints are shown in gray (past) and blue (current) arrows; the next-best viewpoint is shown in red.
    }
    \label{fig:algo_demo}
    \vspace{-18pt}
\end{figure}

However, obtaining only the 3D position of an avocado is insufficient for automated harvesting purposes. 
Avocados are heavier, less uniformly shaped, and have irregular surfaces, making them unsuitable for vacuum or soft-based end-effectors~\cite{zhou2024design}. 
Additionally, they are often cleanly detached from the peduncle to facilitate packaging and avoid puncturing other packaged avocados, which can accelerate rotting. 
Hence, 6D pose information is required so that specific viewpoints (canonical view [CV] and front view [FV]~\cite{zhou2024design}) can be assessed to determine picking points. 

Selecting from a fundamentally small number of picking points can be further hindered by occlusions.  
Several works have focused on fruit pose estimation under occlusion. 
Existing methods can be grouped into three categories: picking point prediction~\cite{yu2019fruit}, fruit shape completion~\cite{marangoz2022fruit, gong2022robotic, pan2023panoptic, yao2024safe}, and viewpoint planning.
The first two focus on predicting the current view using prior knowledge or learning~\cite{gong2022robotic, li2024single}. 
These approaches work for regularly shaped fruit, but are ill-suited for avocados (avocado has more than 500 cultivars~\cite{de2020convolutional}, 
varying drastically in their shape, size, and volume~\cite{mokria2022volume}). 
The latter approach iteratively determines the most informative viewpoint for a sensor to reduce the uncertainty regarding the observed object/area. 
Existing viewpoint planning algorithms have primarily been tested in controlled greenhouses~\cite{menon2023nbv, pan2023panoptic,zaenker2023graph, sun2024efficient}, which differ significantly from avocado cultivation (heavily occluded, unstructured setting).

\begin{figure}[!h]
\vspace{-6pt}
    \centering
    \begin{subfigure}[b]{0.260\textwidth}
        \centering
        \includegraphics[width=\textwidth]{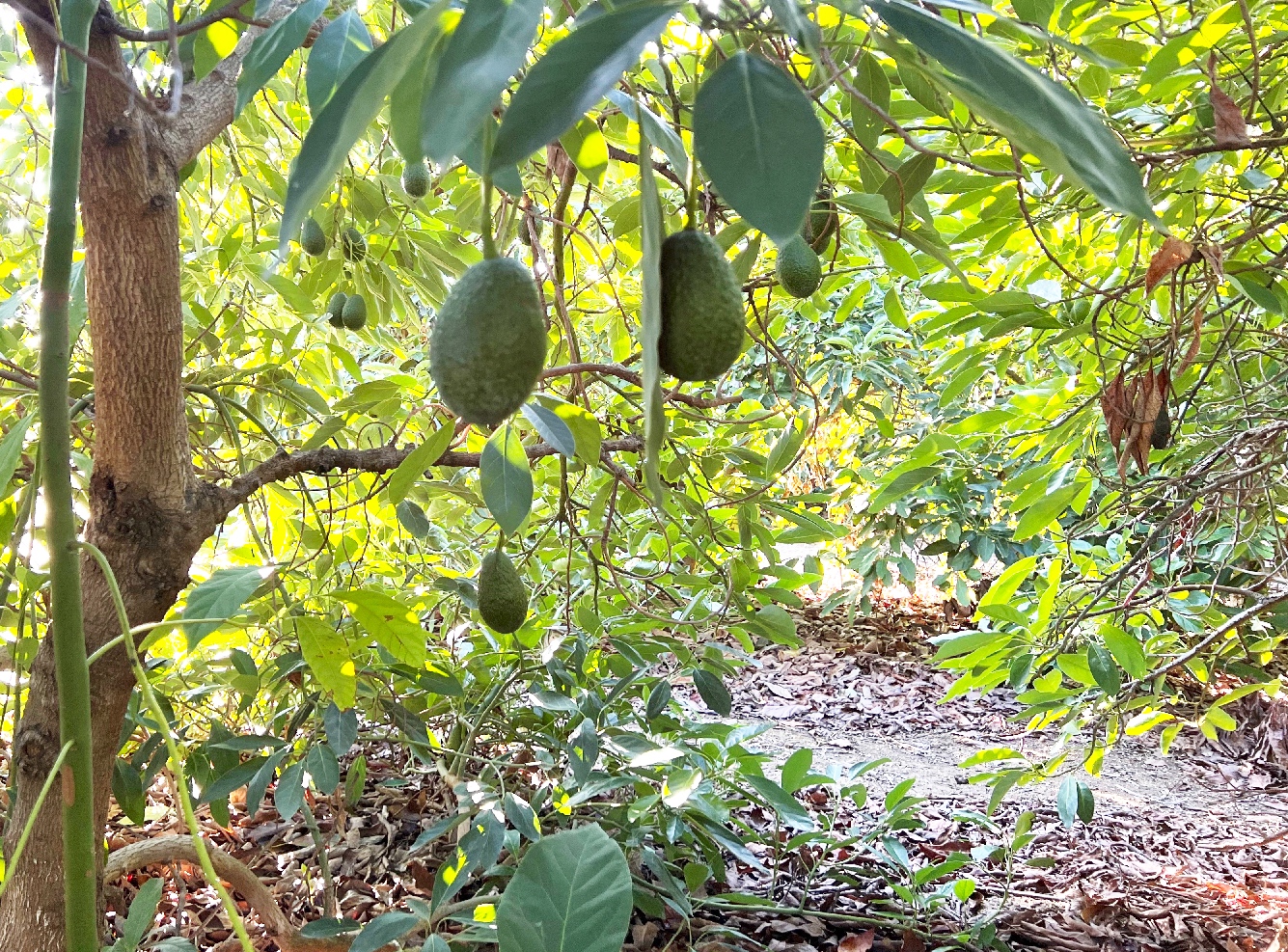}
        \label{fig:avo_env}
    \end{subfigure}
    \hspace{0.0005\textwidth} 
    \begin{subfigure}[b]{0.210\textwidth}
        \centering
        \includegraphics[width=\textwidth]{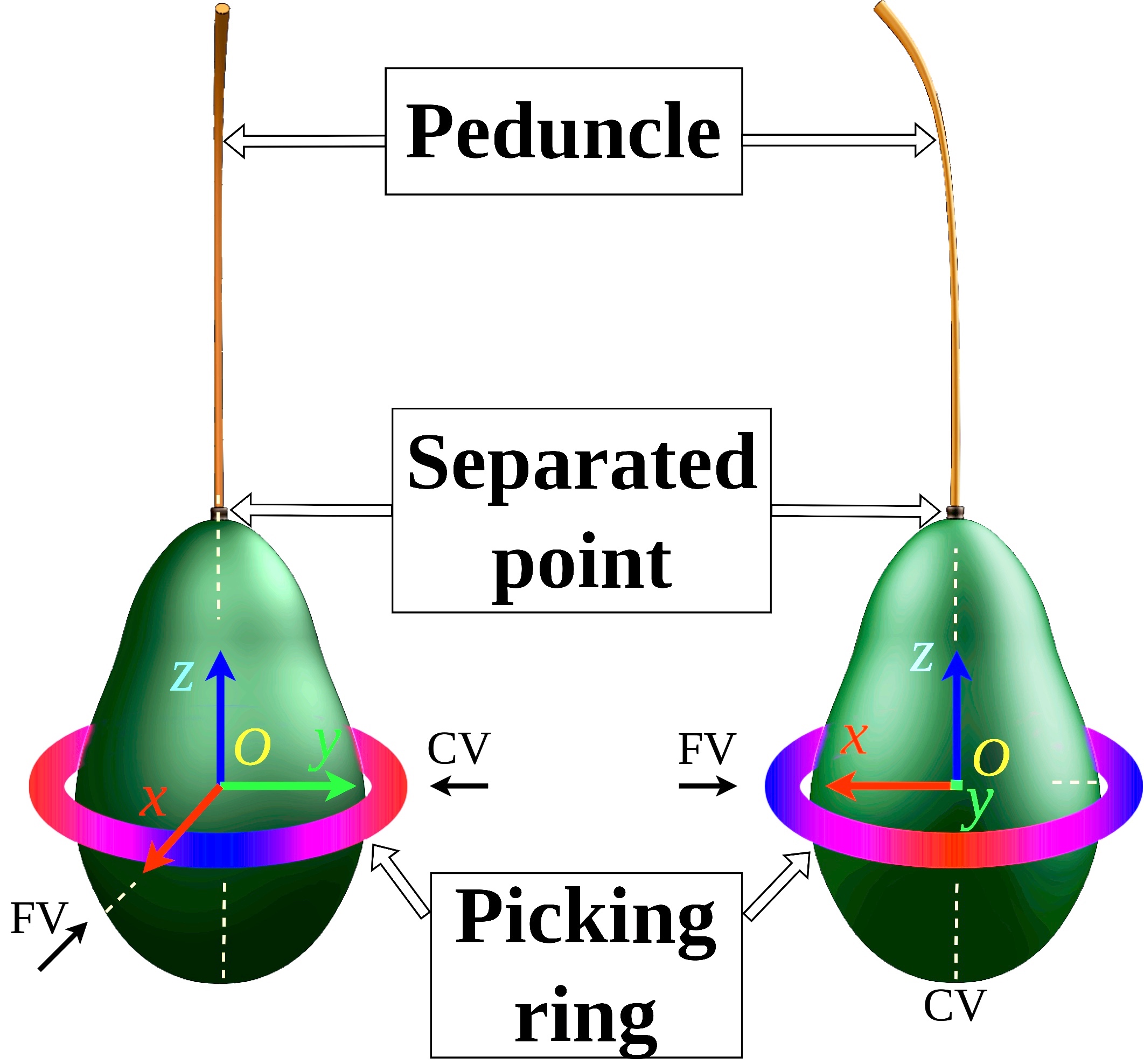}
        \label{fig:picking_ring}
    \end{subfigure}
    \vspace{-26pt}
    \caption{
    Avocado growing environment (left). Fruit coordinate system and picking ring (right).}
    \label{fig:avocado_env_ring}
    \vspace{-12pt}
\end{figure}

In this work, we develop a viewpoint planning algorithm (Fig.~\ref{fig:algo_demo}) to address the unique challenges associated with automated avocado harvesting. 
We employ both geometric and semantic information to sample from a small number of plausible picking points and determine the next-best view that maximizes fruit visibility while adhering to appropriate avocado harvesting viewpoints~\cite{zhou2024design}. 
The paper contributes: 
\begin{itemize}
    \item A geometry-based, semantics-aware viewpoint planning algorithm to tackle occlusion in avocado harvesting. 
    \item  A method to reduce the avocado picking point search space to a 1D ring (Fig.~\ref{fig:avocado_env_ring}) to improve computation.
    \item A new picking score metric definition to directly evaluate the suitability of the 
    selected viewpoints.
\end{itemize}
The ability to handle occlusions can be merged with recent advances in avocado harvesting~\cite{liu2024vision} and field deployment~\cite{liu2024hierarchical} to improve success rates and harvesting efficiency. 

\section{Problem description and set up}
\begin{figure}[!t]
\vspace{6pt}
    \centering
    \begin{subfigure}[b]{0.2\textwidth}
        \centering
        \includegraphics[width=\textwidth]{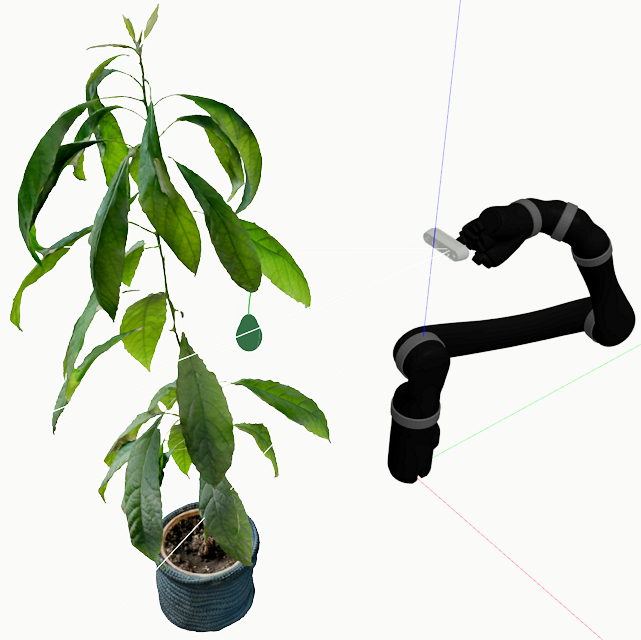}
        \label{fig:cont_sur_curve}
    \end{subfigure}
    \hspace{-5pt} 
    \begin{subfigure}[b]{0.2\textwidth}
        \centering
        \includegraphics[width=\textwidth]{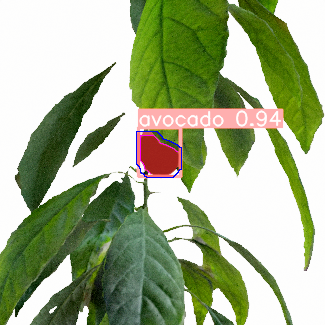}
        \label{fig:sim_env_setup}
    \end{subfigure}
    \vspace{-12pt}
    \caption{Left: Gazebo simulation environment. 
    The avocado was partially visible within the field of view of the camera. 
    Right: The contour (magenta) and surrounding curve (blue) extracted from the avocado mask predicted by the YOLOv8 model~\cite{Jocher_Ultralytics_YOLO_2023}.
    }
    \label{fig:curve_sim_fruit}
    \vspace{-18pt}
\end{figure}

We developed a simulation environment in Gazebo by emulating a potted avocado tree model, a 6-DOF Kinova Jaco2 robot arm, and a RealSense D435 depth camera mounted on the robot end-effector (Fig.~\ref{fig:curve_sim_fruit}). 
We first provide the following definitions. 
\label{section:picking_ring}
\begin{enumerate}
    \item \textit{Picking ring:}  
    Circumferential 1D region in the x-y plane defining the optimal avocado-picking points.
    \item \textit{Discoverability ($s_{dis}$):} Binary metric indicating Object-of-Interest (OOI) visibility; $s_{dis}=1$, if fruit and peduncle are both visible and overlapping or within a 5-pixel proximity and $s_{dis}=0$ otherwise.
    \item \textit{Contour and surrounding curve (C, R):} Fruit contour (C) and dilated surrounding curve (R).
    \item \textit{Occlusion rate ($s_{occ}$):} Proportion of $R$ with depth less than $d_{f\text{min}} +d_{\text{offset}}$, where $d_{fmin}$ is the minimum fruit depth and $d_{\text{offset}}$ is set to $0.06$\;m, i.e. 
    $s_{occ} = {\sum_{p \in R} \mathbb{I}\{d_p < (d_{f\text{min}} + d_{\text{offset}})\}}/|R|$. 
    \item \textit{Picking score ($s_{pick}$):} 
    Success rate of fruit picking from the current view defined as  
    $s_{pick} = s_{dis} * (1-s_{occ})$. A fruit is considered pickable if $s_{pick}>0.9$. 
\end{enumerate}

Given the plant placed within the arm workspace $V\subset\mathbb{R}^3$, the region $V_{\text{ooc}} \subseteq V$represents the area occupied by the plant, whereas $V_{\text{ooi}} \subseteq V_{\text{ooc}}$ 
represents a region that encompasses the avocado, depicted as a green cube in Fig.~\ref{fig:algo_demo}. 
Initially, $V_{\text{ooi}}$ and $V_{\text{ooc}}$ are unknown. 
The planner plans the arm's motion within the free space $V_{\text{free}} \subseteq V$ and 
explores the region $V_{\text{ooi}}$, starting from a pose in which the fruit is partially visible, and finally places the camera in a pose that renders the fruit pickable.

\section{Developed Framework}

\begin{figure}[!h]
\vspace{-6pt}
    \centering
    \includegraphics[width=0.45\textwidth]{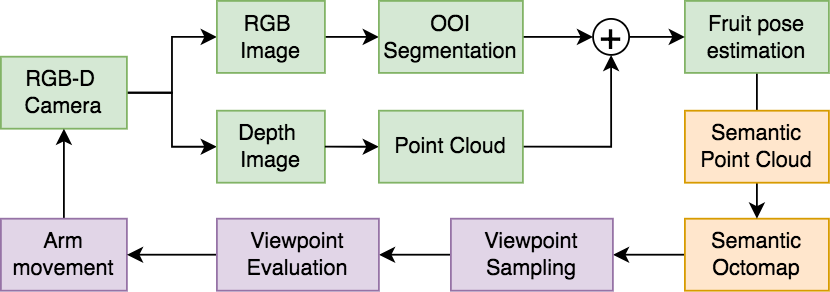}
    \vspace{0pt}
    \caption{System overview. The three modules include sensing (green), 3D representation (yellow), and viewpoint planning (purple).} 
    \label{fig:system_overview}
    \vspace{-18pt}
\end{figure}

We propose a geometry-based, semantics-aware viewpoint planning method (termed GS-NBV), which is an iterative process comprising sensing, 3D representation, and viewpoint planning. Figure~\ref{fig:system_overview} summarizes our framework.

\subsection{Sensing}
\subsubsection{Object Detection} 
The first step is to detect the fruit in the scene. 
We used YOLOv8~\cite{Jocher_Ultralytics_YOLO_2023} to perform instance segmentation on the input image sequences and generate a segmented image with class label, confidence score, and mask. 
The network was fine-tuned to detect avocado and peduncle using a custom dataset of 108 RGB images (640x640 pixels), collected in the simulated environment.\footnote{~Factors like illumination, surface reflections and other sample variations were not considered, but are part of future work.} 
K-fold cross-validation (K=108) was used for training. 

\subsubsection{Fruit Pose Estimation and Update} 
\label{subsubsection:fruit_position}
The fruit pose consists of position and axis, defined as $\boldsymbol{f} = \{\boldsymbol{f}_{pos}, \boldsymbol{f}_{axis}\}$, where $\boldsymbol{f}_{pos} \in V_{\text{ooi}}$.  
The detected fruit and depth values were then converted into a point cloud in the world frame.
To mitigate ill-matching and out-of-range measurement issues, we applied depth clipping at $0.03$\;m (near) and $0.72$\;m (far) to refine the depth map. 
The fruit position was estimated as the mean of the point cloud, with outliers filtered using a one standard deviation threshold. 
The fruit axis was computed as the norm of a vector derived from the highest and lowest fruit points. 
Due to occlusion,  the estimated axis may deviate significantly from the z-axis. 
Here, we elected to bound the angle to a maximum of $30$\;\textdegree{}, as the avocado axis is unlikely to deviate significantly from the z-axis due to the influence of gravity. 
New observations were used to update the fruit position and axis as 
$\boldsymbol{f} = \boldsymbol{f}_{t-1} + k * (\boldsymbol{f}_{t} - \boldsymbol{f}_{t-1})$. 
We set $k=0.7$ to assign a greater weight to the most recent estimates, as newer observations provide better information and are more likely to improve pose estimation.

\subsubsection{Obstacle Points Estimation and Update} 
\label{subsubsection:obstacle_points}
Image-plane points with depth $d_p<d_{fmin}, d_p\in R$ were first transformed into world coordinates and then appended to the list of obstacle points ($P_{occ}$) cumulatively with subsequent observations.

\subsubsection{Picking Score Computation} 
The picking score was computed each time the camera moved to a new viewpoint. 
Planning is concluded once a pickable viewpoint is found.

\subsection{Semantic OctoMap}
Our framework is built on top of the semantic OctoMap method~\cite{burusa2024gradient}.
It is a 4D voxel grid $\mathcal{M} \in \mathbb{R}^{W \times H \times D \times C} $ of width \textit{W}, height \textit{H}, depth \textit{D}, and channels \textit{C}.
To represent the plant, we set \textit{W}, \textit{H}, \textit{D} as $(0.42, 0.45, 0.84)$\;m.
Each voxel has four channels: occupancy probability, semantic class label, semantic probability, and region of interest (ROI) indicator. 
To set the ROI, we defined a $(0.2, 0.2, 0.3)$\;m cube centered at the initial estimated fruit position (shown in green in Fig.~\ref{fig:algo_demo}). 
Although the initial estimation is not 
exact, the ROI is sufficiently large to encompass the entire fruit and peduncle.
Our objective is to explore this ROI to accurately estimate the fruit pose and identify viewpoints that render it pickable. 
The semantic class labels are \textit{background}, \textit{avocado} and \textit{peduncle}, whereas all other settings follow~\cite{burusa2024gradient}.
The semantic OctoMap was updated by combining 
raycasting~\cite{hornung2013octomap} and max-fusion~\cite{semantic_slam} techniques. 

\subsection{Viewpoint Planning: GS-NBV} 
We first define the utility score to evaluate viewpoints and then elaborate on our algorithm (Alg.~\ref{alg:mynbv}). 
\subsubsection{Viewpoint Utility Score}
\label{subsubsection:vp_eva}
The score combines the expected semantic information gain and motion cost. 
The former is defined as $G_{\text{sem}}(\boldsymbol{v}) = \sum_{x \in (X_v \cap V_{\text{ooi}})} T(x)I_{\text{sem}}(x)$ for a given viewpoint $\boldsymbol{v}$~\cite{burusa2024gradient}. 
$X_v$ is the set of all voxels within the field of view, $V_{ooi}$ is the set of ROI voxels, and $T(x)$ is an indicator function that determines the visibility from a viewpoint $\boldsymbol{v}$ to a voxel $x$. 
However, $T(x)$ is conceptualized differently in our approach, whereby the invisible semantic voxels do not contribute to information gain; that is, $G_{\text{sem}}(\boldsymbol{v})=0$ when no semantic voxels are directly visible in the given viewpoint. 
This adjustment enhances the quality of the candidate viewpoints by strictly enforcing visibility constraints.
The semantic information $I_{\text{sem}}(x)$ follows~\cite{burusa2024gradient}, which measures uncertainty via Shannon's entropy; however, we exclude the background class as our focus is on exploring OOIs.
%
The Euclidean distance ($d$) from the current viewpoint was used to select the next-best view. 
The final utility score is $ U_{\text{sem}}(\boldsymbol{v})= G_{\text{sem}}(\boldsymbol{v}) \times e^{-\lambda d}
$, where $\lambda$ controls the motion cost penalty. 
To promote exploration, we set $\lambda = -1$ to move to a more distant viewpoint on the picking ring. 

\subsubsection{Viewpoint Planning Algorithm}
To find the next-best view, the planner first samples the candidate viewpoints. Since the ideal picking points 
lie along the picking ring, we defined a sampling space, $\mathcal{P}$, on a $0.21$\;m radius circle in 3D space 
centered at $\boldsymbol{f}_{pos}$ 
(this radius can be adjusted, but cannot be less than the sensor's effective measuring distance).
We constrained the sampling space to a $270$\;\textdegree{} arc, as the rear portion of the tree is harder to reach with a robotic arm (this limitation can be removed with an aerial platform).
This defines
two boundary points $p_{b}^{1}$ and $p_{b}^{2}$.

\begin{algorithm}
\caption{GS-NBV planning}
\label{alg:mynbv}
\begin{algorithmic}[1]
\Statex \textbf{input}: initial pose $\boldsymbol{v_{init}}$, No. of candidate viewpoints $n$
\Statex \textbf{output}: poseList   // all planned poses
\State $t \gets 0$;
\State $\boldsymbol{v} \gets \boldsymbol{v_{init}}$;
\State $planningFlag \gets True$;
\While {$t \leq N_{max}$ \textbf{and} $planningFlag$}
    \State $s_{occ}, P_{occ}, 
    \boldsymbol{f}, 
    D, Sems = sensing(\boldsymbol{v})$;
    \State $s_{pick} = s_{dis} * (1-s_{occ})$; 
    \State $\mathcal{M} = updateSemanticOctomap(D, Sems, \boldsymbol{v})$;
    \If {$s_{pick} > 0.9$}
        \State $planningFlag \gets False$;
        \State exit() // planning succeed;
    \EndIf
    \State $V_{vps} \gets uniformAdaptiveSampling(\boldsymbol{f}
    , P_{occ})$;
    \State $\boldsymbol{v_{nbv}} = \arg\max_{v \in V_{vp}}(score(\mathcal{M}, \boldsymbol{v}))$;
    \State Arm.move($\boldsymbol{v_{nbv}}$);
    
\EndWhile
\end{algorithmic}
\end{algorithm}

\begin{figure}[!t]
\vspace{0pt}
    \centering
    \includegraphics[width=0.49\textwidth]{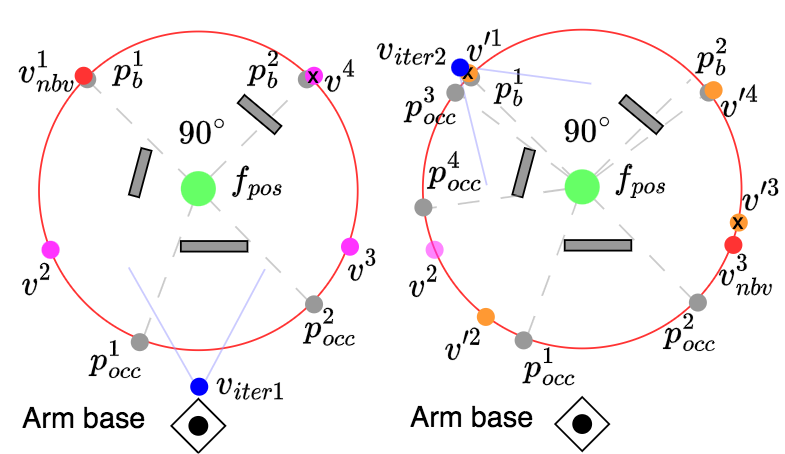}
    \vspace{-12pt}
    \caption{Exemplary GS-NBV 
    algorithm operation. The black dot represents the fixed arm base. The blue dot indicates the camera's current viewpoint, and the occluded fruit is discoverable within the camera's field of view (blue lines). The green dot represents the estimated avocado position $\boldsymbol{f}_{pos}$, and the red circle represents the picking ring. For brevity, we assume $\boldsymbol{f}_{pos}$ is fixed and $\boldsymbol{f}_{axis}$ always points outward, perpendicular to the plane. 
    The gray bars represent background objects in the environment (such as leaves or other tree parts), and the gray dots indicate boundary points. 
    Initially, the camera is positioned at $v_{iter1}$. 
    In the $1^{st}$ iteration, four points on the picking ring (purple dots) are selected. $v_4$ is discarded (marked with a cross), and $v_1$ is selected as the next best viewpoint (red dot). 
    In the $2^{nd}$ iteration, the camera moves to $v_1$, and the fruit pose and boundary points are updated. 
    Four new points are selected (orange dots); $v'_1$ and $v'_3$ are discarded, and $v_3$ is the next-best viewpoint.}
    \label{fig:vp_sampling_algo}
    \vspace{-16pt}
\end{figure}

With reference to Fig.~\ref{fig:vp_sampling_algo}, the algorithm starts with extracting the leftmost and rightmost obstacle points $p_{occ}^{1}$ and $p_{occ}^2$ from the current viewpoint $v_{iter1}$. 
This reduces $\mathcal{P}$ to $\overset{\huge\frown}{p_{b}^{1}p_{occ}^{1}}\cup\overset{\huge\frown}{p_{occ}^{2}p_{b}^{2}}$.
The iterative decrease in $\mathcal{P}$ permits subsampling of viewpoints. 
Herein, we selected four equidistant viewpoints $v^1$, $v^2$, $v^3$ and $v^4$ from the updated $\mathcal{P}$. 
Instead of random sampling~\cite{yi2024view,zaenker2021viewpoint}, our approach leverages geometric information to reduce the computational cost of the raycasting algorithm while ensuring adequate spatial coverage of the tree from various angles. 
Each viewpoint is represented by a 7-element vector (i.e. position and orientation in quaternion form). 
The orientation is generated from the directional vector produced by the position of the viewpoint relative to $\boldsymbol{f}_{pos}$.  
Newly sampled viewpoints that fell within a $0.1$\;m radius of previously sampled viewpoints were removed because adjacent viewpoints generally do not provide sufficiently new information. 
This operation is often called \textit{viewpoint dissimilarity filtering}~\cite{menon2023nbv}.  
Hence, $v_4$ was discarded in the first iteration, and the update of the boundary point $p_b^2$ was performed in the second iteration (right panel of Fig.~\ref{fig:vp_sampling_algo}).  
The remaining candidate viewpoints are ranked in descending order based on their total utility scores and stored in a Viewpoint Queue, $V_{vps}$. 
Lastly, the next-best viewpoint is chosen as $\boldsymbol{v}_{\text{nbv}} = \arg\max_{\boldsymbol{v} \in V_{vps}} U_{\text{sem}}$.
The first iteration resulted in the sensor moving to $v_1$.

As new observations are acquired, both the fruit pose and semantic OctoMap are updated. 
A new picking ring is estimated based on the updated fruit pose. 
Subsequently, we project previous boundary points onto this new circle. This process ensures that the sampling space continually decreases, thereby enhancing the viewpoint sampling efficiency. 
In the second iteration, two more boundary points, $p^3_{occ}$ and $p^4_{occ}$ are incorporated as the camera detects more obstacle points.
Hence, $\mathcal{P}$ was updated to 
$\overset{\huge\frown}{p_{b}^{1}p_{occ}^{3 }}\cup 
\overset{\huge\frown}{p_{occ}^{4}p_{occ}^{1}} \cup \overset{\huge\frown}{p_{occ}^{2}p_{b}^{2}}$. 
Uniform equidistant sampling therefore yielded new viewpoints: $v'^1$, $v'^2$, $v'^3$ and $v'^4$. 
However, $v'^1$ and $v'^3$ were excluded because they fell within $0.1$\;m radius of previously sampled viewpoints. 
Newly selected viewpoints and their utility scores are saved to $V_{vps}$ in descending order.

Despite new iterations introducing additional observations and further exploring the map, only the motion cost part for viewpoints already in the queue is updated. 
Expected semantic information gains are not re-calculated. 
This utility update strategy optimizes efficiency. 
It reduces the computational load by avoiding raycasting updates on the entire queue and biases the selection of informative newer viewpoints when available; otherwise, it defaults to selecting previously assessed viewpoints that may also carry new information in the newer iterations. 
Based on these, $v^3$, a previously selected viewpoint with the highest utility in the queue, was chosen as the next-best view in the second iteration. 
If the queue is empty, a new set of viewpoints is re-sampled following the same rules outlined above.
This process continues iteratively until the fruit is pickable.

\section{Experimentation and Results}
\subsection{Experimental Setup}

\begin{figure}[!t]
\vspace{6pt}
    \centering
    \includegraphics[width=0.49\textwidth]{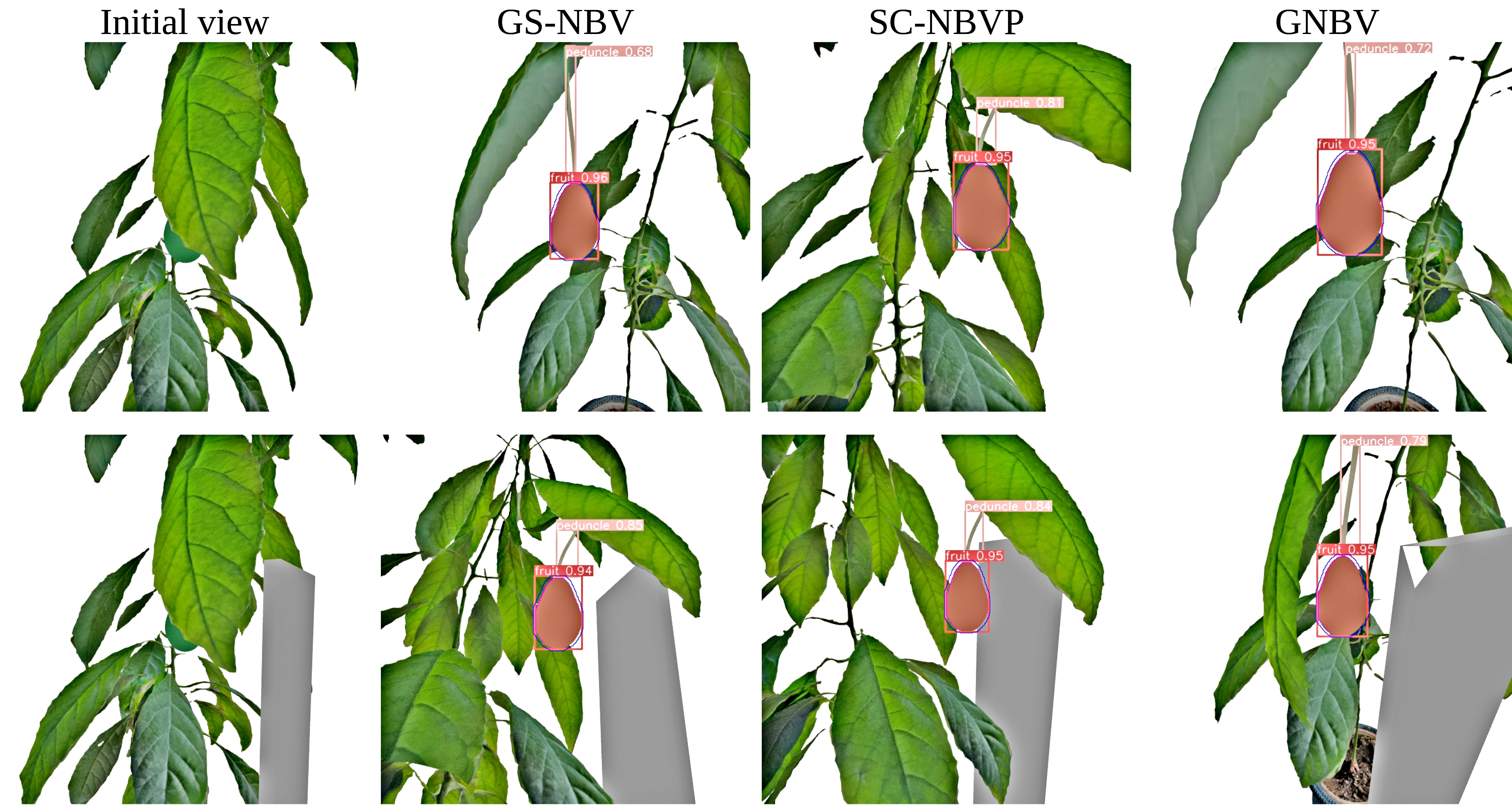}
    \vspace{-12pt}
    \caption{Initial view and the best final view provided by each algorithm in Group 1 (Top) and Group 2 (Bottom) case studies.
    }
    \label{fig:init_end_plot}
    \vspace{-3pt}
\end{figure}

We considered two case studies in a static environment (Fig.~\ref{fig:init_end_plot}): Group 1, an initial view with fruit visibility at 27.98\%; Group 2, the same setup but with a board placed to the right side of the plant that completely blocks the visibility of viewpoints from the right side. 
A simulated avocado was placed vertically and occluded by leaves. 

\begin{figure}[!t]
\vspace{6pt}
    \centering
    \includegraphics[width=0.47\textwidth]{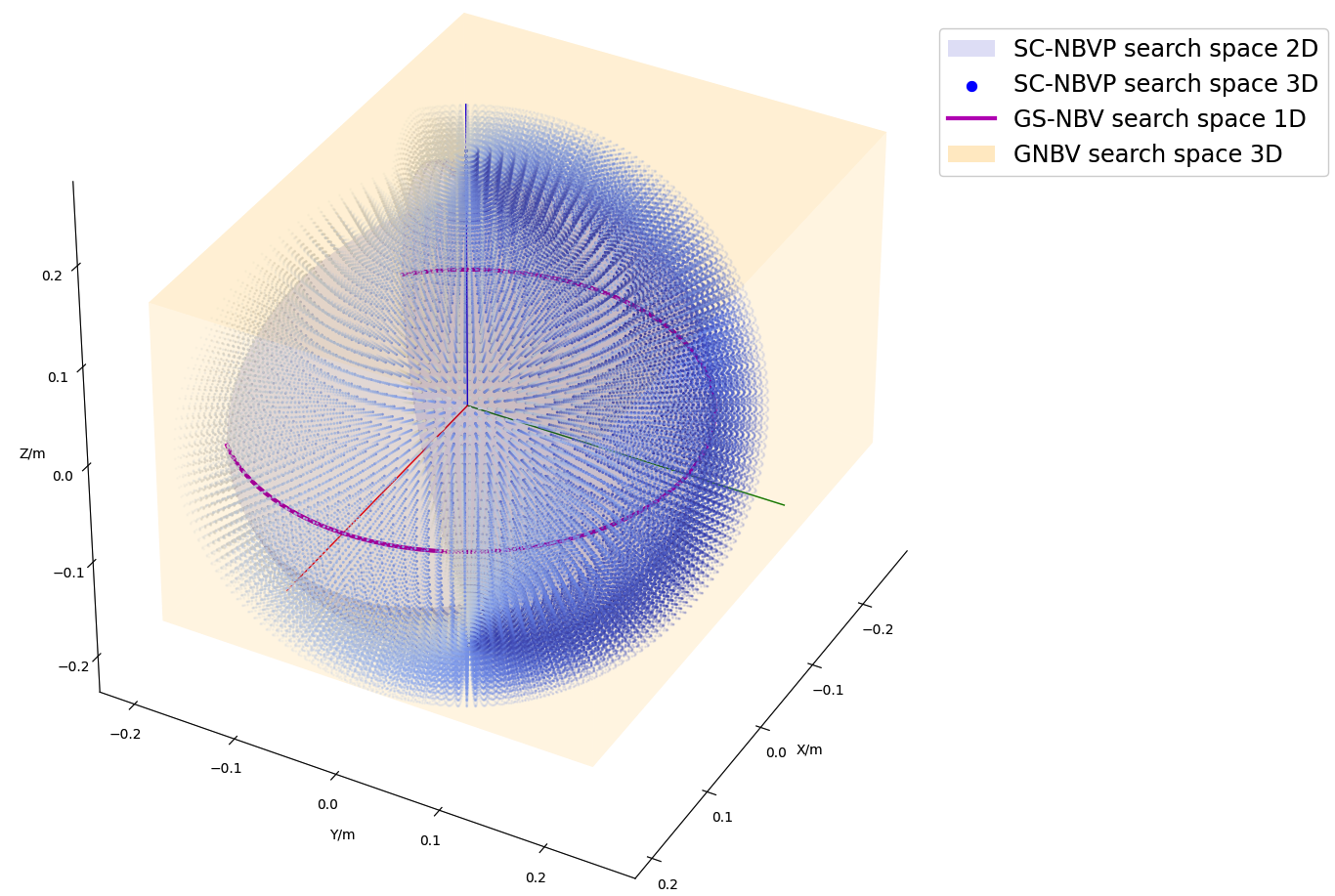}
    \vspace{-2pt}
    \caption{Search spaces for all planners. (Best viewed in color.)
 }
    \label{fig:search_space_compare}
    \vspace{-16pt}
\end{figure}

We tested our method against the sampling-based SC-NBVP method~\cite{yi2024view} and the gradient-based GradientNBV (GNBV) approach~\cite{burusa2024gradient}, which have shown good performance in the context of local planning in similar occlusion settings. 
We used the reported parameter values and further fine-tuned them for our setup (Table~\ref{table:sota_para_compare}). 
For fairness, the ROI was kept constant for all planners (green cube in Fig.~\ref{fig:algo_demo}). 
To accelerate SC-NBVP, we reduced its sampling space from 3D to 2D. 
The viewpoint sampling spaces for all planners are shown in Fig.~\ref{fig:search_space_compare}. 
All arm motions were performed using MoveIt~\cite{moveit}.

\begin{table}[!t]
\vspace{0pt}
\centering
\caption{Parameters of All Planning Methods}
\vspace{-3pt}
\label{table:sota_para_compare}
\begin{tabular}{lccc}
\toprule
\textbf{Parameter}&\textbf{GS-NBV}&\textbf{SC-NBVP} & \textbf{GNBV} \\
\midrule
RRT step size \(L\)                 & n.a        & 0.1m      & n.a                  \\
RRT max.nodes \(N_{\text{max}}\)    & n.a          & 10        & n.a                  \\
Sampling dissimilarity     & 0.1m           & n.a       & n.a                \\
Number of candidate views \(n\)     & 4           & 4                  & 1        \\
The motion cost coefficient \(\lambda\)     & -1        & 1           & n.a         \\
Spherical sampling radius \(R\)
& n.a          & 0.1m             & n.a          \\
Octomap resolution \(r\)    &\multicolumn{3}{c}{0.003m}\\
Discrete point spacing \(\sigma\)   &\multicolumn{3}{c}{0.003m}  \\
Camera minimum depth measure
&\multicolumn{3}{c}{0.03m}   \\
Camera maximum depth measure 
&\multicolumn{3}{c}{0.72m}\\
Arm maximum reach&\multicolumn{3}{c}{0.60m}\\
\(R_{\text{min}}\) &0.21m &0.21m& n.a  \\
\(R_{\text{max}}\) &0.21m &0.21m & n.a \\
Octomap max-fusion coefficient &0.9 & n.a & 1.0 \\
Step size $\alpha$ &n.a& n.a & 0.65 \\
Picking score threshold &\multicolumn{3}{c}{0.90}\\
\bottomrule
\end{tabular}
\vspace{-16pt}
\end{table}

All planners were run on a laptop equipped with an Intel i7-9750H CPU running at 2.60 GHz, 32 GB of RAM, and an NVIDIA RTX 2070 GPU with 8 GB memory.  
Each planner was tested in 10 trials under identical conditions for each group of trials. 
Each planning task ended either when a view rendering the fruit pickable was achieved or when a maximum of 10 planning iterations was reached.

\subsection{Evaluation Metrics}
The following metrics were used for evaluation.  
\emph{Successful rate} is the percentage of successful pickable viewpoints out of 10 trials. 
\emph{Unoccluded rate} is defined as $1-s_{occ}$;  higher values indicate higher probability of successful fruit picking. 
A successful planning process requires the fruit to be fully discoverable with an unoccluded rate exceeding a preset threshold of $0.9$. 
\emph{Fruit pose error} includes average position and orientation error of the estimated avocado center and axis from the ground truth values (set herein to $[0.01, -0.36, 0.559]$\;m and $[0, 0, 1]$, respectively).  
\emph{Number of planning iterations} is the average number of planning iterations a planner required to complete the task. 
\emph{Time consumption} is the average time a planner required to complete a single viewpoint planning iteration.

\subsection{Planning Time Results and Analysis}
Table~\ref{table:eva_nbv_time_compare} shows comparative results of the planners' average time consumption. Three main components are presented: viewpoint sampling, viewpoint evaluation, and NBV computation.  
Bold font represents the best performance. 
Note that the time consumption of GNBV is zero, as the gradient-based method does not sample viewpoints. 
Due to more severe occlusion conditions, it is necessary to map the entire tree rather than only the ROI. This increased the time required for a single raycasting call in SC-NBVP compared to the original work ($0.136-0.162$\;sec). 
The number of candidate views are 4, 4 and 1 for our planner, SC-NBVP, and GNBV, respectively, with the viewpoint evaluation process requiring $1.473$, $3.765$, and $0.130$\;sec, respectively. 
In the NBV computation process, our method only selects the viewpoint with the highest utility; thus, the time consumption is negligible. 
SC-NBVP selects the viewpoint with the highest utility via randomized sampling (by the RRT algorithm), resulting in a slightly longer computation time. 
GNBV requires the gradient of the information gain function to be computed for the entire image, thereby requiring the most time. 
The entire NBV planning process takes $1.576$, $4.054$, and $0.461$\;sec, respectively. 
In summary, GNBV is faster owing to the use of gradient computation, but among the two sampling-based methods, our approach is $2.5\times$ faster than SC-NBVP. 

\subsection{Planner Performance Results and Analysis}
Table~\ref{table:success_rate} presents a comparative analysis of planner performance across experiment groups, with the best results 
highlighted in bold. 
Metrics other than the success rate were based only on successful trials. 
SC-NBVP's success rate improved from 30\% to 60\% in the more challenging Group 2. 
This is because viewpoints from the right side 
generate high SC scores~\cite{yi2024view}, making them less likely to be selected. 
This is equivalent to reducing the sampling space, thus improving the success rates and the No. of planning iterations. 
For the same reason, our planner also takes fewer iterations in Group 2 (on average 1.2 times) than in Group 1 (1.5 times). 
In both experimental groups, GNBV showed a similar performance in terms of success rate and No. of planning iterations.

\begin{table}[!t]
\vspace{6pt}
\centering
\caption{Planning Time Comparison }
\vspace{-3pt}
\label{table:eva_nbv_time_compare}
\begin{tabular}{lccc}
\toprule
\textbf{Sub-process} [s]&\textbf{GS-NBV}&\textbf{SC-NBVP} & \textbf{GNBV} \\
\midrule
Viewpoints sampling&0.017        & 0.040     & \textbf{0}               \\
\midrule
Single raycasting call&0.199     & 0.645
& \textbf{0.130}\\
Viewpoints evaluation   &1.473       & 3.765   &    \textbf{0.130}         \\
\midrule
Compute next-best viewpoint   & \textbf{0.000}       & 0.072    & 0.328               \\
\midrule
NBV planning & 1.576 & 4.054     &\textbf{0.461}\\  
\bottomrule
\end{tabular}
\end{table}

\begin{table}[!t]
\vspace{0pt}
\centering
\caption{Planner Performance Comparison}
\vspace{-3pt}
\label{table:success_rate}


\resizebox{0.49\textwidth}{!}{ 
\begin{tabular}{ccccc} 
\toprule
Metric & Group & GS-NBV & SC-NBVP& GNBV \\
\midrule
Success rate [\%] & \multirow{4}{*}{1} & \textbf{100\%} & 30\% & 60\% \\
Fruit position error [m] & & \textbf{0.04$\pm0.00$}
 & 0.07$\pm0.08$ &0.15$\pm0.03$ \\
 Fruit axis error [$\circ$] &  & 22.63$\pm7.37$&\textbf{22.49}$\pm8.24$&30.00$\pm0.00$\\
No. planning iterations& & \textbf{1.5} & 6&2.7\\
\addlinespace 
Success Rate [\%] & \multirow{4}{*}{2} & \textbf{100\%} & 60\% & 50\% \\
Fruit position error [m] &  &\textbf{0.02$\pm0.01$}&$0.09\pm0.09$&$0.13\pm0.06$\\
Fruit axis error [$\circ$] &  & $30.00\pm0.00$&\textbf{24.09$\pm9.90$}&$28.82\pm2.36$\\
No. planning iterations& & \textbf{1.2}& 4.5&2.6\\
\bottomrule
\end{tabular}
}
\vspace{-6pt}
\end{table}

\begin{figure}[!t]
\vspace{6pt}
    \centering
    \includegraphics[width=0.48\textwidth]{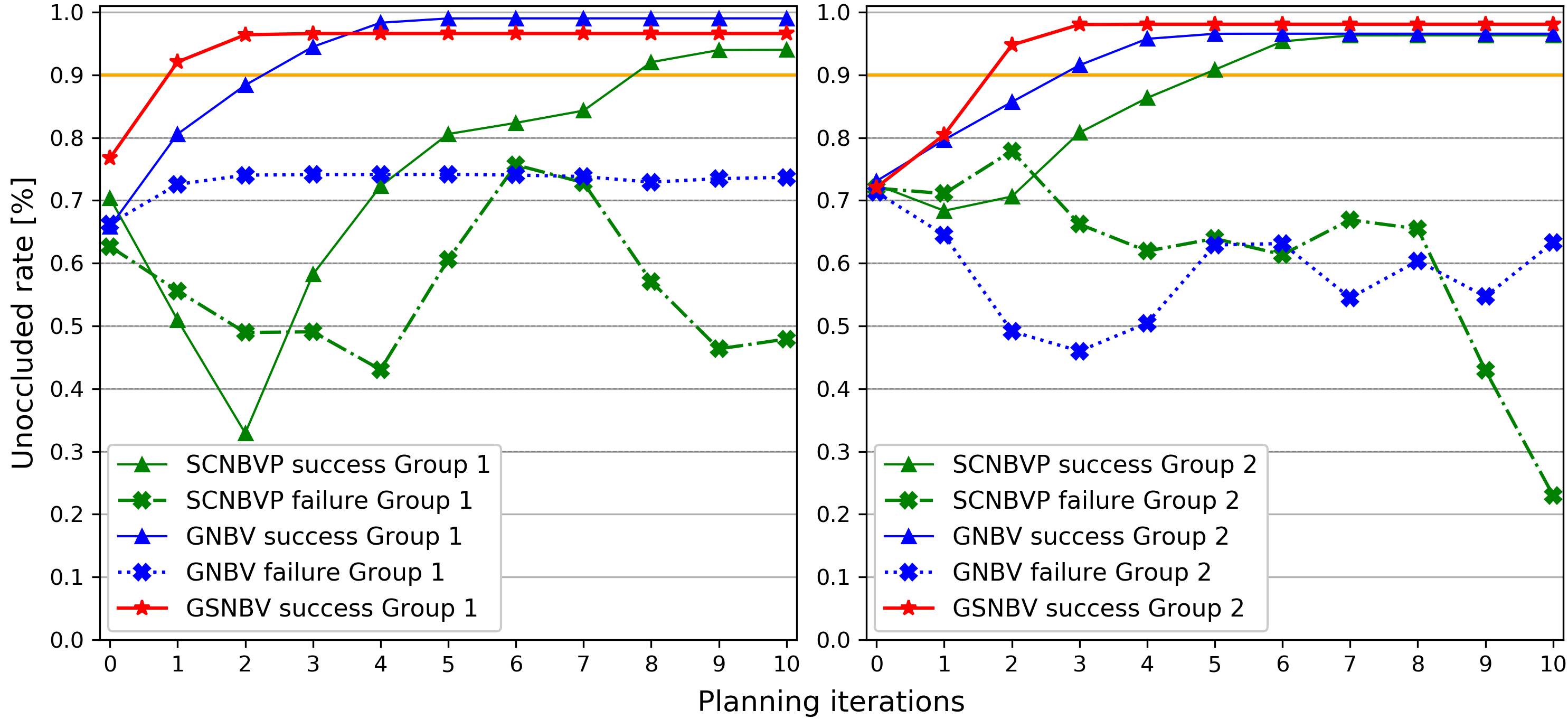}
    \vspace{-12pt}
    \caption{
    Avg. unoccluded rate plot.
    Solid and dashed lines indicate the results of successful and failed attempts, respectively. 
    The orange line (y = 0.9) marks the picking threshold. Once the fruit is 
    fully 
    discoverable, 
    planning stops, and the curve flattens.
    }
    \label{fig:result_plot}
    \vspace{-18pt}
\end{figure}

Figure~\ref{fig:result_plot} presents the average unoccluded rate curves for successful and failed attempts. 
Due to sampling randomness, the \textit{SC-NBVP failure curve} fluctuates as the planning iterations increase and does not converge. 
On the other hand, the \textit{GNBV failure curve} converges to a local maximum and does not increase with more planning iterations. 
In contrast, our method consistently achieved a 100\% success rate in both groups and identified a pickable view within two planning iterations. 
Further, Fig.~\ref{fig:init_end_plot} shows qualitative results of the planners. 
The best final views captured by each planner in both settings are shown.
In Group 1, all planners identified an ideal view. 
In Group 2, SC-NBVP found an optimal view, whereas our planner's final view was from below the fruit, and GNBV's was from above. 
However, GNBV's final view requires the arm to navigate through a narrow gap between the leaves and the board, which may be risky in practice.

\subsection{Fruit Pose Estimation Results and Analysis}
Our method accurately estimated the fruit position with the lowest mean error and standard deviation in both groups owing to the stable planner performance and equally consistent sampling distance on the picking ring. 
In contrast, the randomness in viewpoint sampling and RRT algorithm in SC-NBVP planner led to relatively high mean position errors and the highest standard deviation. 
GNBV had a smaller standard deviation than SC-NBVP, as it typically operates within a localized area. 
However, this also led to the largest mean position error among the three planners. 
%
In both groups, none of the planners accurately estimated the fruit axis, which poses additional challenges for the final picking step (beyond the scope of this paper but planned for future work). 
However, this limitation does not hinder their ability to identify viable picking views under heavy occlusion.

\section{Conclusions and Future Research}
This work introduced a geometry-driven, semantic-aware viewpoint planning algorithm for avocado harvesting under occlusion.
By constraining the search to a 1D picking ring and introducing a new picking score, our method significantly improves planning efficiency and success.
Simulation results show a 100\% success rate with minimal iterations, outperforming two state-of-the-art methods (max 60\%).
These results highlight the effectiveness and robustness of our algorithm, which shows promise for improving avocado harvesting efficiency.  
Future work will explore the effect of varying the method's hyperparameters, focus on sim-to-real domain transfer to enable physical field testing, and explore global planning for multi-tree harvesting. 

\bibliographystyle{IEEEtran}
\bibliography{references}

\end{document}